\documentclass[10pt,twocolumn,letterpaper]{article}

\usepackage{cvpr}              %

\usepackage{booktabs}
\usepackage{lipsum}
\usepackage{subcaption}
\usepackage{stmaryrd}
\usepackage{pifont}%
\usepackage{graphicx}
\usepackage{algorithm}
\usepackage{algpseudocode}
\usepackage{setspace}
\usepackage{balance}

\newcommand{\cmark}{\ding{51}}%

\newcommand{\MethodName}{MIRANDA }

\definecolor{cvprblue}{rgb}{0.21,0.49,0.74}
\usepackage[pagebackref,breaklinks,colorlinks,allcolors=cvprblue]{hyperref}

\def\paperID{65} %
\def\confName{EarthVision}
\def\confYear{2026}

\title{MIRANDA: \underline{MI}d-feature \underline{RAN}k-adversarial \underline{D}omain \underline{A}daptation 
\\ toward climate change-robust ecological forecasting with deep learning}
\author{Yuchang Jiang\\
{\tt\small yuchang.jiang@uzh.ch}
\and
Jan Dirk Wegner\\
{\tt\small jandirk.wegner@uzh.ch}
\\\vspace{8pt}
DM$^3$L, UZH, Zurich, Switzerland
\and
Vivien Sainte Fare Garnot\\
{\tt\small vivien.saintefaregarnot@uzh.ch}
}

\begin{document}
\maketitle
\begin{abstract}

Plant phenology modelling aims to predict the timing of seasonal phases, such as leaf-out or flowering, from meteorological time series. Reliable predictions are crucial for anticipating ecosystem responses to climate change. While phenology modelling has traditionally relied on mechanistic approaches, deep learning methods have recently been proposed as flexible, data-driven alternatives 
with often superior performance. 
However, mechanistic models tend to outperform deep networks when data distribution shifts are induced by climate change.
Domain Adaptation (DA) techniques could help address this limitation. Yet, unlike standard DA settings, climate change induces a temporal continuum of domains and involves both a covariate and label shift, with warmer records and earlier start of spring. To tackle this challenge, we introduce \underline{Mi}d-feature \underline{Ran}k-adversarial \underline{D}omain \underline{A}daptation (MIRANDA). Whereas conventional adversarial methods enforce domain invariance on final latent representations, an approach that does not explicitly address label shift, we apply adversarial regularization to intermediate features. Moreover, instead of a binary domain-classification objective, we employ a rank-based objective that enforces year-invariance in the learned meteorological representations.
On a country-scale dataset spanning 70 years and comprising 67,800 phenological observations of $5$ tree species, we demonstrate that, unlike conventional DA approaches, \MethodName improves robustness to climatic distribution shifts and narrows the performance gap with mechanistic models. \href{https://github.com/SherryJYC/MIRANDA}{https://github.com/SherryJYC/MIRANDA}

\end{abstract}
    
\section{Introduction}
\label{sec:intro}

\begin{figure}
    \centering
    \includegraphics[width=\linewidth]{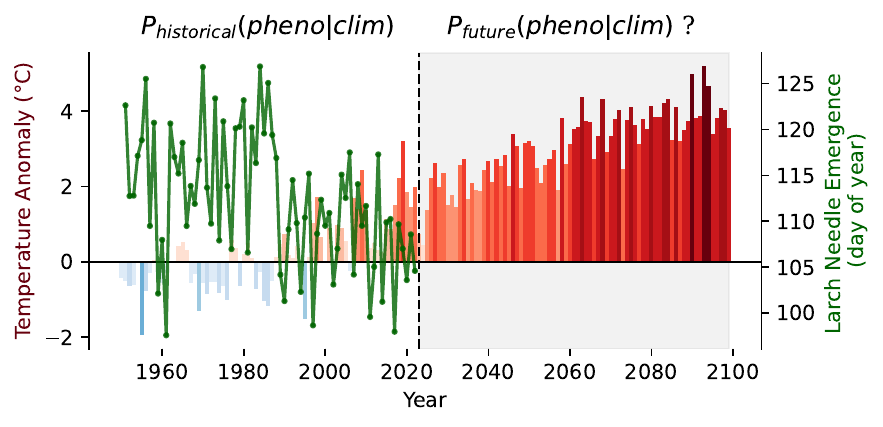}
    \caption{\textbf{Climate change-robust machine learning.} Deep learning models trained on historical climatic conditions face a distribution shift when applied to future climate projections. In this paper we consider models that predict plant phenological dates from meteorological time series and devise a domain adaptation method to enhance their robustness to such shifts. (Temperature data: CMIP6 CNRM-CM6-1 SSP2-4.5 \cite{eyring2016cmip6, voldoire2019CNRMCM6}, phenological data: Swiss Phenology Network \cite{SPN}) }
    \label{fig:teaser}
\end{figure}

Plant phenology, the study of seasonal vegetation cycles, focuses on the timing of events such as leaf emergence, flowering, and fruit maturation, which are driven by environmental and climatic factors \cite{schwartz2003phenology}. Climate change has already induced widespread shifts in phenological timing, providing some of the most visible biological evidence of ongoing global change \cite{cleland2007shifting}. These shifts have far-reaching ecological consequences, affecting trophic interactions \cite{plantbird, plantpolinator} and ecosystem functioning, and they feed back to the climate system through their influence on biomass production, surface albedo, and carbon and water fluxes \cite{piao2019climatefeedback2, richardson2010influence}. Beyond ecology and climate science, phenology also underpins many Earth observation applications, including crop type mapping, biomass estimation, and species distribution modelling \cite{hoppe2024transferability, croci2025assessing, chen2023evaluating, ponti2023importancephenoSDM}. Reliable phenology models are therefore essential for anticipating the impacts of future climate change on ecosystems, land-climate interactions, and remote sensing applications.

\begin{figure}
    \centering
    \includegraphics[width=\linewidth, trim=0cm 24.7cm 8.5cm 0cm, clip]{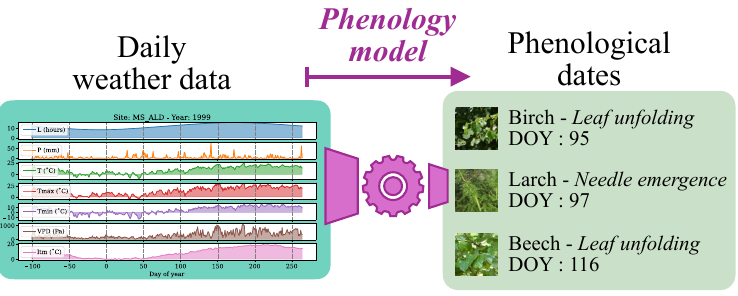}
    \caption{\textbf{Phenology modeling (conceptual illustration).} A phenology model takes a set of climatic and environmental covariates and predicts phenological dates. In this paper we consider models operating on daily meteorological time series to predict the date of $50\%$ leaf development of $5$ different tree species. }
    \label{fig:pheno-model}
\end{figure}

In this work, we consider the task of climate-phenology modelling, which consists of predicting species-specific phenological dates from one year of daily meteorological time series (Fig. \ref{fig:pheno-model}). Such regression models are typically trained on phenological observations derived from near-surface and satellite remote sensing \cite{young2025phenocamv3, caparros2021land}, or  citizen-science networks \cite{SPN}. Traditionally, phenology modelling has relied on mechanistic, hypothesis-driven models that encode predefined physiological responses to temperature and day length \cite{hufkens2018phenor}. More recently, deep learning approaches have been proposed, achieving superior performance under stationary climatic conditions \cite{garnot2025phenoformer}. Unlike process-based models, data-driven approaches can flexibly incorporate a broader range of predictors and scale more readily to large spatial extents or diverse species sets \cite{garnot2025phenoformer}.
However, deep models are prone to learning correlations that are specific to the historical climate distribution on which they are trained. As a result, they often underperform mechanistic models when evaluated under climatic conditions that deviate substantially from training data \cite{asse2020process, garnot2025phenoformer}. This limitation is critical for ecological forecasting, since future climate projections include environmental conditions that may lie outside the historical training distribution \cite{deepa2024climatextremes}, as illustrated in Fig.~\ref{fig:teaser}.

Our goal in this paper is to improve the robustness of deep phenology models under such distribution shifts. We frame this challenge as an Unsupervised Domain Adaptation (UDA) problem. Historical observations define a \emph{source} domain, where both meteorological inputs and phenological labels are available. Future climate simulations define a \emph{target} domain, for which meteorological inputs are accessible but phenological labels are unavailable \cite{eyring2016cmip6}. This setting naturally enables unsupervised adaptation using unlabeled future climate data. However, phenology forecasting presents challenges that are not well addressed by existing UDA methods. Phenology modelling under climate change is subject to both a covariate shift, with a trend of warming climatic records, and a label shift with a trend of advancing date of spring phenological events. While some UDA methods address this combined shift with distribution re-weighting strategies, these frameworks are designed for classification settings and are ill-suited here. Furthermore, climate change induces a temporal continuum of domains rather than two clearly separated domains. Applying existing methods could thus result in data invariance that are harmful to the model's robustness. 
To address these challenges, we introduce \underline{MI}d-feature \underline{RAN}k-adversarial \underline{D}omain \underline{A}daptation (\MethodName), to the best of our knowledge, this is the first domain adaptation method designed explicitly for phenology modelling under climate change. To address the combined covariate and label shift, we apply an adversarial regularizer on the intermediate features of a transformer model that enforce domain-invariance. Applying this regularizer here instead of on the final latent features allows for domain-specific variations in the last part of the model, which we further expand with a hybrid normalization layer. Secondly, instead of a binary domain classification objective for the adversarial network, we implement a rank-based loss that pushes the main network to learn year-invariant intermediate features. We emulate the UDA setting using a 70-year historical dataset from a country-scale citizen-science phenology network with $67,800$ different phenological observation. We construct three different dataset splits mimicking three different intensities of distribution shift. Through extensive experiments, we show that, unlike conventional UDA methods, \MethodName  improves performance under strong distribution shifts and  narrows the performance gap between deep learning and process-based mechanistic models in extrapolative settings.

\section{Related Work}

\paragraph{Phenology modelling} 
Process-based models implement clear causal mechanisms linking climatic drivers to plant phenological response \cite{caffarra2011modelling, hufkens2018phenor}. Such models, like the M1 model \cite{blumel2012M1}, typically have high interpretability, small number of parameters, and robust performance. However, they hinge on theoretical knowledge of the pathway that links a driver to plant phenology. This requires extensive controlled experiments in growth chambers \citep{hanninen2019experiments}. 
In contrast, data-driven approaches can learn predictive relationships directly from observational data, enabling the incorporation of diverse  drivers without requiring explicit mechanistic assumptions \cite{lee2022RFKoreaColouration, rodriguez2015RFrsPheno, dai2019gbmchina}.  
Recent deep learning architectures such as PhenoFormer \cite{garnot2025phenoformer} have been developed for plant phenology and showed significantly superior performance than process-based models when tested in stationary climatic conditions. However, when evaluated under shifted climate conditions, mimicking global warming, the M1 process-based models still outperformed PhenoFormer. 
This challenge is critical in realistic forecasting scenarios where models are trained on historical observations, and deployed on simulated future meteorological data to anticipate phenological trends for the next century \cite{eyring2016cmip6, pau2011predicting}. 
The versatility and scalability of data-driven methods could benefit phenological forecasting, enabling, for example, the inclusion of drivers like water availability which are currently not  embedded in process models \cite{piao2019climatefeedback2}. However, this climatic robustness challenge needs to be addressed first.

\paragraph{Domain adaptation}
Unsupervised domain adaptation (UDA) aims to transfer predictive models from a labeled source domain to an unlabeled target domain under distribution shift, and has become a central topic in machine learning when annotation is scarce or expensive \cite{wilson2020surveyuda, liu2022udareview}. A common principle of many UDA methods is to use the unlabeled target samples to encourage the extraction of \emph{domain-invariant} representations. Indeed, if the representations are domain-invariant, the mapping from these representations to the label space will likely remain valid in the target domain. There are broadly three types of methods to achieve this. \emph{Divergence-based} approaches regularize model training with a loss term which aims at minimizing the statistical divergence between the distributions of source and target representations, such as Maximum Mean Discrepancy \cite{long2015learningMMD} or second order statistics \cite{CORAL}. \emph{Adversarial approaches} implement an adversarial training objective where a discriminator network tries to predict the domain of latent representations while the main network tries to make this task impossible, effectively erasing domain-specific patterns in the representations \cite{ajakan2014DANN, tzeng2017ADDA}. \emph{Normalization-based} approaches rely on the statistics of normalization layers to make the representations domain-agnostic \cite{romijnders2019DANL}. Beyond these methods focusing on domain-invariance, some works incorporate self-supervised objectives aiming at learning features that are more robust to domain shifts \cite{sun2019unsupervised, saito2020universal}. 
Although much of the UDA literature originates from computer vision, domain adaptation techniques have also been explored for time series data, which introduce additional challenges \cite{shi2022deep,he2023raincoat, gackstetter2025raincoatSITS}. %
However, most existing approaches focus on classification tasks or multi-source adaptation, and are thus not directly suited for phenology modeling.
Moreover, many UDA methods assume covariate shift, where only the input distribution changes. In phenology forecasting, climatic changes induce not only covariate shift but also observable label shift in phenological dates (see Fig. \ref{fig:shifts}). While some recent works address label-related shifts \cite{he2023raincoat, gackstetter2025raincoatSITS}, they do so in classification settings. In contrast, we propose a domain adaptation framework tailored to time-series regression under combined covariate and label shift, specifically designed for phenology modelling, a regression problem.

\begin{figure*}[t!]
    \centering
    \includegraphics[width=\linewidth, trim=0cm 7.1cm 0cm 0cm, clip]{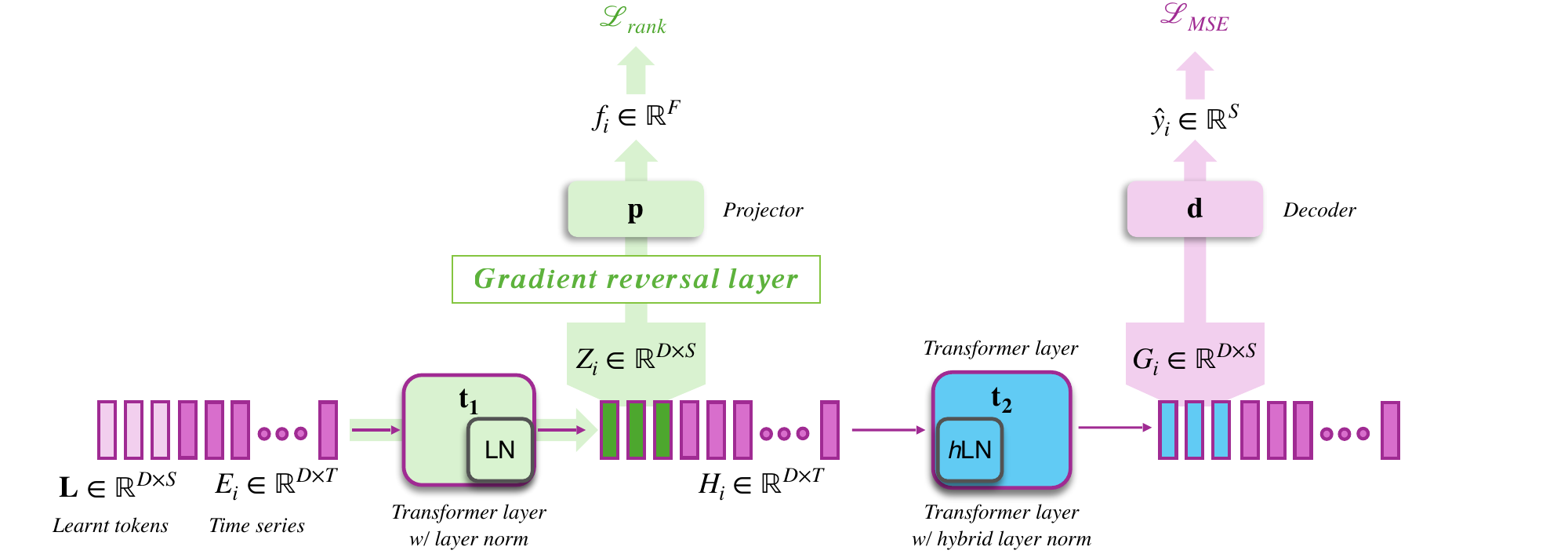}
    \caption{\textbf{Method overview}. 
    Our framework addresses domain shifts in phenology modelling via two key components: rank-based adversarial training on intermediate features and hybrid layer normalization.
    Purple elements correspond to the {\color{Orchid}\textbf{main phenology prediction pathway}}: the learnable tokens $\mathbf{L}$ are concatenated with the input time series $E_i$ and processed by two transformer encoder layers ($\mathbf{t_1}$ and $\mathbf{t_2}$). The resulting global embeddings of the learnable tokens, denoted as $G_i$, are then fed into the decoder $\mathbf{d}$ to predict the target date using the regression loss $\mathcal{L}_{\mathrm{MSE}}$.
    We apply {\color{YellowGreen}\textbf{rank-based adversarial learning}} on the mid-level features $Z_i$ (green), where a discriminator $\mathbf{p}$ is trained with a ranking loss $\mathcal{L} _{\mathrm{rank}}$ through a gradient reversal layer to encourage domain-invariant mid-level features.
    Meanwhile, we replace the standard layer normalization in $t_2$ with our {\color{cyan}\textbf{hybrid layer normalization}} (blue), which preserves domain-dependent variations in high-level representations.
    }
    \label{fig:method}
\end{figure*}

\section{Methods}

\subsection{Problem statement}

\paragraph{Phenology modeling}
We consider the task of phenology modelling from meteorological time series. A model $\mathbf{F}_\theta$, parametrized by $\theta$, operates on a meteorological time series $X_i \in \mathbb{R}^{C \times T}$, with $C$ the number of climate variables and $T$ the length of the temporal sequence to predict a set of phenological dates $\hat{Y_i} \in \mathbb{R}^S$ with $S$ the number of different tree species under consideration (across our experiments $T=365$, $C=7$, and $S=5$). Formally, this is a multi-variate regression problem from multi-temporal inputs. The dataset is organized in \emph{site-years}: the same sites are monitored over multiple years and sample $i$ correspond to the meteorological conditions and phenological observations of site $s_i$ during year $g_i$.

\paragraph{Unsupervised domain adaptation} We cast the task of future phenology projection as an unsupervised domain adaptation problem. In this setting, we make use of a \emph{source} dataset $\mathcal{D}_{\sigma}$ and a \emph{target} dataset $\mathcal{D}_{\tau}$. The source dataset is fully labeled: $\mathcal{D}_{\sigma} = \{(X_i, Y_i) \in \mathbb{R}^{C \times T} \times \mathbb{R}^S\}_{i=1}^{N_{\sigma}}$, with $N_{\sigma}$ the number of training samples and $Y_i$ the observed phenophase dates for sample $i$. In contrast, only unlabeled inputs are available in the target domain  $\mathcal{D}_{\tau} = \{(X_i) \in \mathbb{R}^{C \times T} \}_{i=1}^{N_{\tau}}$, with $N_\tau$ the number of target samples.

\subsection{Model architecture} We use PhenoFormer \cite{garnot2025phenoformer}, an established architecture for the task at hand,  as backbone for our method. PhenoFormer is a lightweight architecture composed of a shared linear encoder, a transformer encoder layer with learnt tokens, and a linear decoder. Our only modification is to use two consecutive transformer layers instead of one, leaving all other hyperparameter unchanged. 

\begin{align}
    X_i & \xrightarrow{\mathbf{e}} E_i  \in \mathbb{R}^{D \times T} \label{eq:lin-enc}\\
    [E_i+ p_i , \mathbf{L}] &  \xrightarrow{\mathbf{t_1}} [H_i , Z_i] \in \mathbb{R}^{D \times (T+S)} \label{eq:trf1}\\
    [H_i , Z_i] &  \xrightarrow{\mathbf{t_2}} [\tilde{H}_i , G_i] \in \mathbb{R}^{D \times (T+S)}  \label{eq:trf2}\\
    G_i  & \xrightarrow{\mathbf{d}} \hat{Y_i} \in \mathbb{R}^S  \label{eq:dec}
\end{align}

A given input multi-variate time series $X_i$ is firstly embedded by a linear layer $\mathbf{e}$, applied in parallel to all time points, resulting in a sequence of embedding vectors $E_i$ of dimension $D$, with $D$ a hyperparameter. Temporal positional encodings $p_i$ are added to these embeddings, and learnt tokens $\mathbf{L}$ are appended to the sequence. We use one learnt token per predicted species so $\mathbf{L} \in \mathbb{R}^{D \times S}$. The resulting sequence is processed by two successive transformer encoder layers $\mathbf{t}_1$ and $\mathbf{t}_2$. Lastly, the output embeddings $G_i \in \mathbb{R}^{D \times S}$ corresponding to the learnt tokens' positions are used as global embedding for each species and decoded into a date prediction with a linear decoder layer $\mathbf{d}$. See \citet{garnot2025phenoformer} for more details on this architecture.

\subsection{\MethodName}
We introduce \MethodName a domain adaptation method specifically designed for climate change robustness in phenology modelling. As shown in Fig. \ref{fig:method}, our method relies on 1. a rank-based adversarial objective that is better suited to the continuously changing climatic shift,  2. the adversarial regularization of the intermediate (instead of late) features, which only enforces domain-invariance in the earlier part of the network, 3. a hybrid normalization layer in the last transformer layer of the architecture, which further promotes better extrapolation.

\paragraph{Rank-based adversarial objective} Classical adversarial domain adaptation formulates the discriminator objective as binary domain classification: the discriminator predicts whether a feature originates from the source or the target domain, while the feature extractor is trained to confuse it. This formulation implicitly assumes two clearly separated and discrete domains. However, as visible in Fig \ref{fig:teaser}, in climate-phenology forecasting, domain shift occurs progressively over time, resulting in a continuum of intermediate regimes rather than a sharp source-target dichotomy. Moreover, what constitutes a domain-shift might be different for different tree species. Collapsing this structure into a binary objective discards valuable information about the relative proximity between climatic conditions over time. To better capture this gradual shift, we replace the binary adversarial objective with a Rank-N-Contrast loss \citep{zha2023rank}. Instead of enforcing indiscriminate domain invariance, this objective encourages features to be more similar when they are coming from closer years, and less similar otherwise. This ranking-based formulation naturally accommodates continuous domain shifts and provides a smoother regularization signal than binary discrimination.

Within a mini-batch of size $N$, each pair of feature vectors $(f_i, f_j)$, is encouraged by a contrastive term, to be closer than a set of negative pairs. The negative set $S_{i,j}$ is defined as the set of samples $k$ that are more distant in time than samples $i$ and $j$: 

\begin{equation}
S_{i,j}=\{f_k \mid k\neq i,\ |g_i-g_k|\ge |g_i-g_j|\},
\end{equation}

where $g_i$ is the year of observation of sample $i$.
We use the cosine similarity $\mathrm{sim}(\cdot,\cdot)$ to measure \emph{closeness} in feature space such that the contrastive term for sample $i$ is :
\begin{equation}
    l_{\mathrm{rank}}^{(i)}=\frac{1}{N}\sum_{j=1,\,j\neq i}^{N}
    -\log\frac{\exp\left(\mathrm{sim}(f_i,f_j)/\tau\right)}
    {\sum_{f_k\in S_{i,j}}\exp\left(\mathrm{sim}(f_i,f_k)/\tau\right)} \:,
\end{equation}
with $\tau$ is a temperature hyperparameter. The final adversarial loss $\mathcal{L}_{\mathrm{rank}}$ is then:
\begin{equation}
    \mathcal{L}_{\mathrm{rank}} = \frac{1}{N}\sum^N_{i=1} l_{\mathrm{rank}}^{(i)} \:.
\end{equation}

\paragraph{Mid-feature regularization}
Unlike prior adversarial adaptation methods such as DANN \citep{ajakan2014DANN}, which apply domain discrimination on late features ($G_i$), we perform adversarial alignment on the mid-level representations ($Z_i$) of the learnt tokens. Indeed, if the late features were perfectly aligned across domains, the distributions of the predictions would also be aligned across domains, and thus not be able to track the label shift. Instead, we propose to only enforce domain invariance in the intermediate representations of our model. 
Specifically, we flatten $Z_i \in \mathbb{R}^{D \times S}$ into a vector and process it with a multi-layer perceptron $\mathbf{p}$ to obtain an embedding $f_i \in \mathbb{R}^{F}$, with $F$ a hyperparameter. 
The discriminator $\mathbf{p}$ is trained to minimize $\mathcal{L}_{rank}$, while a gradient reversal layer (GRL) from DANN~\citep{ajakan2014DANN} is inserted between $\mathbf{p}$ and the feature extractor $\mathbf{t_1}$.
As a result, $\mathbf{t_1}$ is optimized adversarially to produce embeddings that cannot be separated according to their year order, leading to domain-invariant mid-level representations. 
The overall training objective is 
\begin{equation}
\mathcal{L} = \mathcal{L}_{\mathrm{MSE}} + \lambda \mathcal{L}_{\mathrm{rank}},
\end{equation}
where $\mathcal{L}_{\mathrm{MSE}}$ is the supervised regression loss computed on labeled source samples,
\begin{equation}
\mathcal{L}_{\mathrm{MSE}} = \|y - \hat{y}\|_2^2,
\end{equation}
and $\mathcal{L}_{\mathrm{rank}}$ is applied to both source and target samples since it does not rely on phenology labels. Each mini-batch is composed of an equal number of source and target samples. The overall training framework is shown in Algorithm \ref{alg:overall_training}.

\begin{algorithm}
\setstretch{0.6}
\small{
\caption{Overall training framework}\label{alg:overall_training}
\begin{algorithmic}
\Require Source batch $\mathcal{B}_s=\{(x_i^s,y_i^s)\}$, target batch $\mathcal{B}_t=\{x_j^t\}$ with $|\mathcal{B}_s|=|\mathcal{B}_t|$
\For{each training iteration}
    \[
    Z^s=\mathbf{t_1}(x^s), \qquad Z^t=\mathbf{t_1}(x^t)
    \]
    \State 1. Regression on source samples
    \[
    \hat{y}^s=\mathbf{d}(\mathbf{t_2}(Z^s)) 
    \]
    \State 2. Mid-feature projection and gradient reversal
    \[
    f^s=\mathbf{p}(\mathrm{\textbf{GRL}}(Z^s)), \qquad
    f^t=\mathbf{p}(\mathrm{\textbf{GRL}}(Z^t))
    \]
    \State 3. Optimization of regression and adversarial training
    \[
    \mathcal{L}_{\mathrm{MSE}}=\|y^s-\hat{y}^s\|_2^2
    \]
    \[
    \mathcal{L}_{\mathrm{rank}}=\mathcal{L}_{\mathrm{rank}}(f^s) + \mathcal{L}_{\mathrm{rank}}(f^t)
    \]
    \[
    \mathcal{L}=\mathcal{L}_{\mathrm{MSE}}+\lambda \mathcal{L}_{\mathrm{rank}}
    \]
\EndFor
\end{algorithmic}
}
\end{algorithm}

\paragraph{Hybrid layer normalization}
While rank-based adversarial training enforces domain invariance at the mid-level representations, the final task-specific features should retain sufficient flexibility to accommodate label shifts in phenological dates.
To enhance flexibility, we propose \emph{hybrid layer normalization} ($h\mathrm{LN}$), which replaces the standard layer normalization in the second transformer layer $t_2$. Our approach is inspired by~\cite{romijnders2019DANL}.  Given a generic input feature vector $x \in \mathbb{R}^{D}$, standard layer normalization computes the mean and standard deviation from the current sample:
\begin{equation}
\mu(x)=\frac{1}{D}\sum_{d=1}^{D} x_d,
\qquad
\sigma(x)=\sqrt{\frac{1}{D}\sum_{d=1}^{D}(x_d-\mu(x))^2}
\end{equation}
and applies normalization as
\begin{equation}
\mathrm{LN}(x)=\gamma\frac{x-\mu(x)}{\sigma(x)}+\beta,
\end{equation}
where $\gamma$ and $\beta$ are learnable affine parameters. Our $h\mathrm{LN}$ follows the same standard functioning in the source domain and differs in the target domain. $h\mathrm{LN}$ maintains \emph{source-domain population statistics} during training in a running-average manner similar to batch normalization. At inference time on the target domain, $h\mathrm{LN}$ uses these running statistics to normalize target feature vectors.

For each training batch, we compute the layer normalization statistics $\mu(x)$ and $\sigma(x)$ of the source domain samples and update the running estimates as:
\begin{equation}
\mu_s \leftarrow (1-m)\mu_s + m\,\mu(x), \qquad
\sigma_s \leftarrow (1-m)\sigma_s + m\,\sigma(x),
\end{equation}
where $m$ is the momentum coefficient.
During inference, $h\mathrm{LN}$ normalizes target features using the source-domain running statistics $(\mu_s, \sigma_s)$:
\begin{equation}
h\mathrm{LN}(x)=\gamma\frac{x-\mu_s}{\sigma_s}+\beta,
\end{equation}
This domain-specific normalization, enables the late features $G_i$ of the target domain to have a different range than those of the source domain.

\section{Experiments}

\subsection{Data and domain definition}

\paragraph{PhenoFormer dataset.}
We use the dataset introduced in~\cite{garnot2025phenoformer}, based on phenological records from the Swiss Phenology Network (SPN)~\citep{SPN}. 
The dataset comprises 67,800 observations collected from 175 sites across Switzerland over up to 70 years.  
We focus on a subset of five tree species that offers the longest record: horse chestnut, European beech, European larch, common spruce, and hazel. We predict the date of needle emergence for European larch and common spruce, and the date of leaf unfolding for the other species. 
Each sample includes daily meteorological time series with $365$ data points for seven variables between day of year $-100$ and $265$ of the year at hand. The meteorological variables cover temperature, precipitation, pressure, and  day length.

\begin{figure*}[t]
    \centering
    \begin{subfigure}{0.28\linewidth}
     \includegraphics[width=\linewidth]{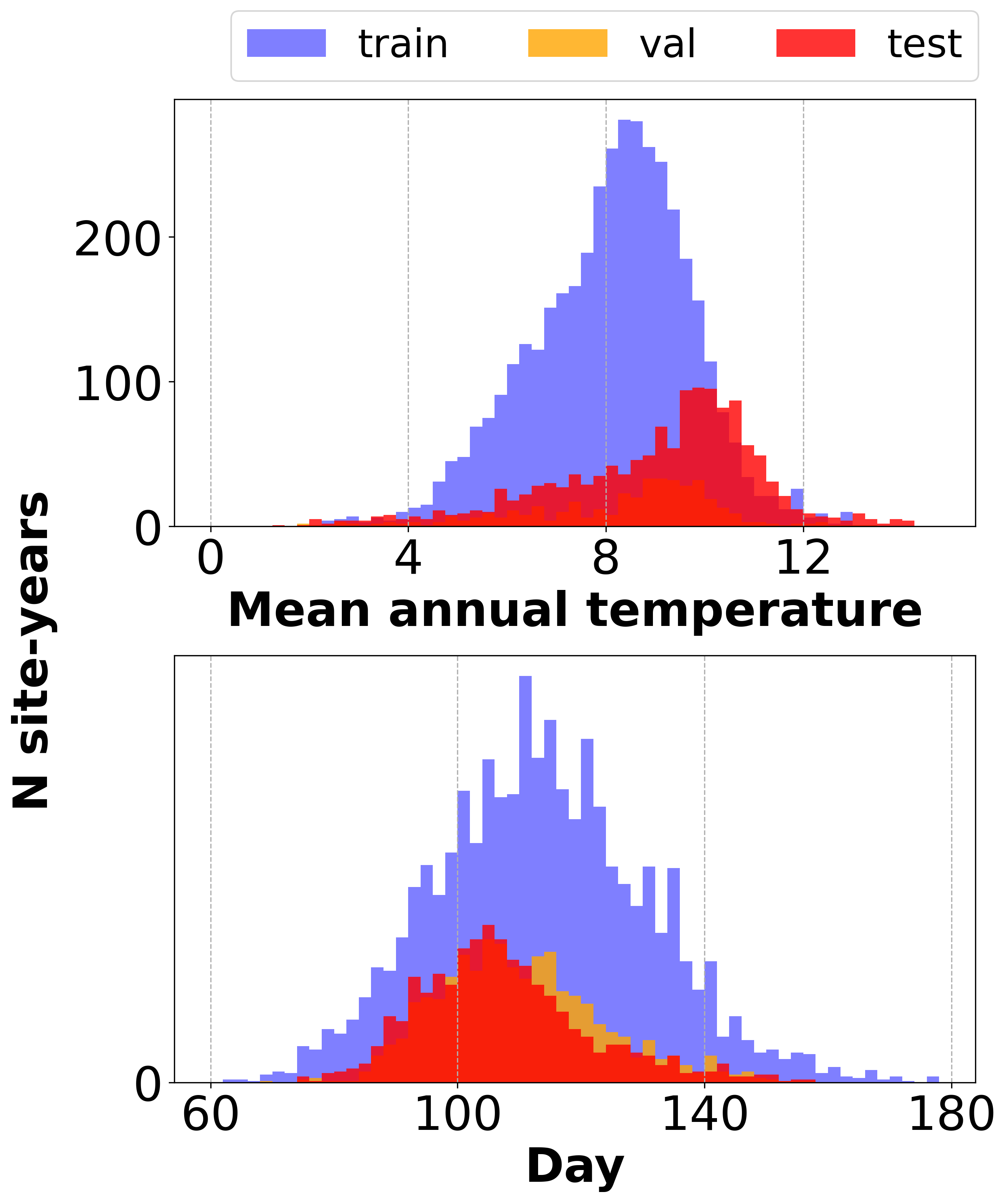}
        \label{fig:shifts-1}
        \caption{Chronological split. \\
        $\Delta T = 0.90^{\circ}C$, 
        $\Delta Date = -6.22\ day$
        }
    \end{subfigure}
    \hfill
    \begin{subfigure}{0.28\linewidth}
        \includegraphics[width=\linewidth]{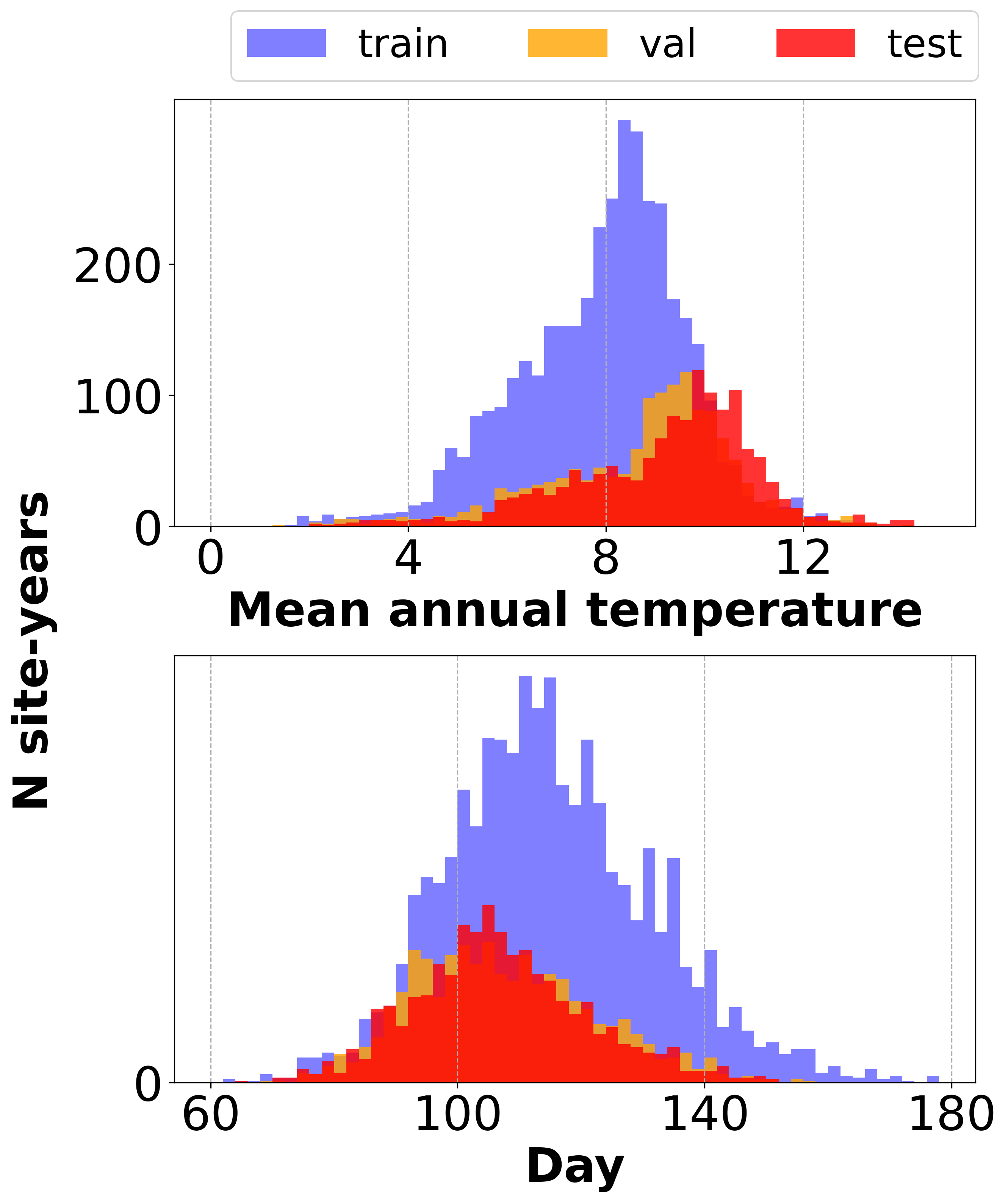}
        \label{fig:shifts-2}
        \caption{Annual temperature split \\
        $\Delta T = 1.23^{\circ}C$, 
        $\Delta Date = -7.92\ day$
        }
    \end{subfigure}
    \hfill
    \begin{subfigure}{0.28\linewidth}
        \includegraphics[width=\linewidth]{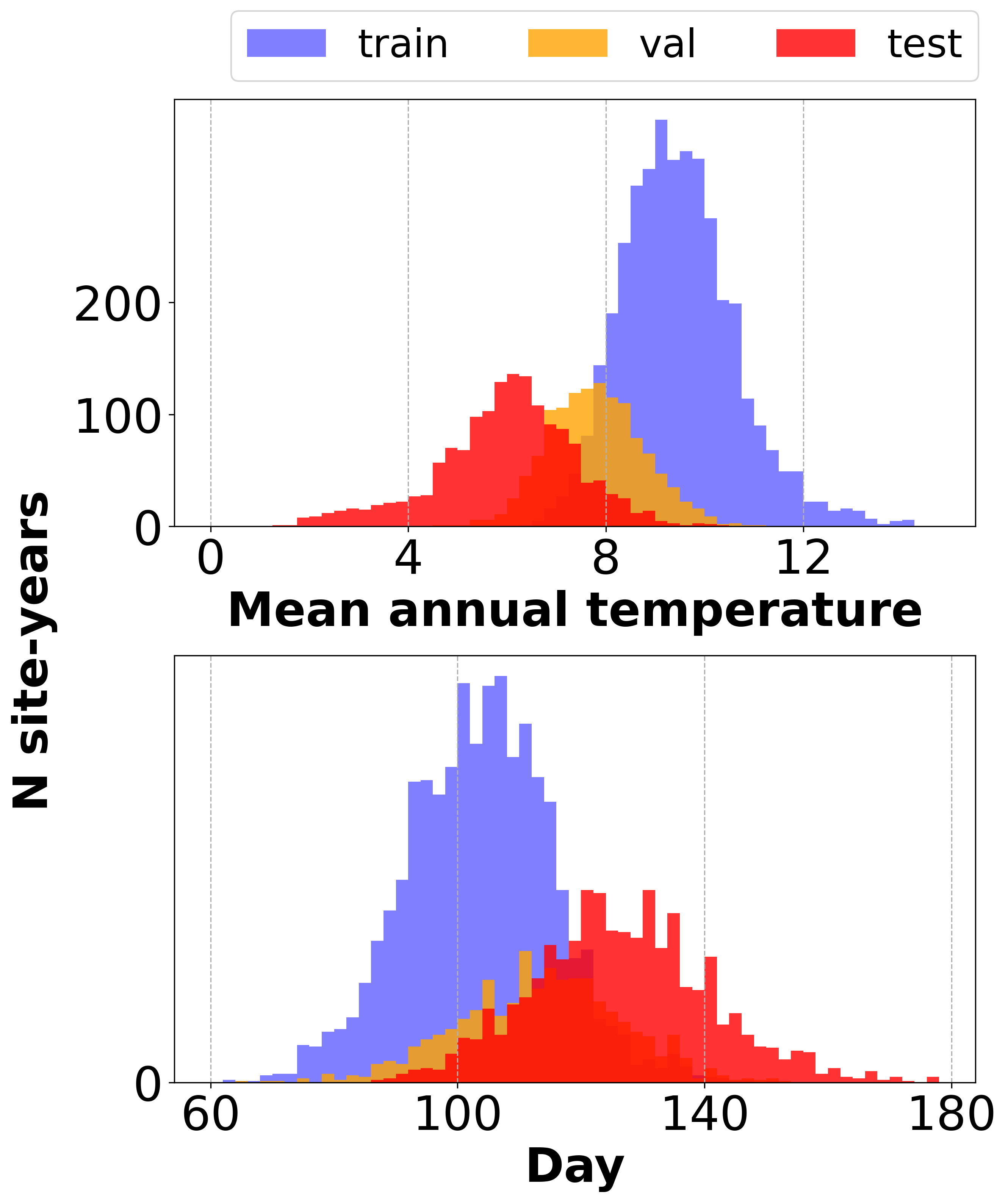}
        \label{fig:shifts-3}
        \caption{Elevation split \\
        $\Delta T = -3.51^{\circ}C$,
        $\Delta Date = 22.96\ day$
        }
    \end{subfigure}
    \caption{\textbf{Domain distribution shifts.} We show the histograms of input (top) and predicted (bottom) variables on the train, validation, and test sets for each of the three dataset splits. We plot the histogram of annual temperature in the top row to represent the climatic distribution, and the date of Larch needle emergence in the bottom row to illustrate the shift in predicted variables.  }
    \label{fig:shifts}
    
\end{figure*}

\paragraph{Dataset splits}

We construct three dataset splits. For all splits, the training set is treated as the source domain, while the validation and test sets are treated as target domains.

\begin{itemize}
    \item \textbf{Chronological} 
    First, we partition the dataset chronologically: observations from 1951--2002 are used for training, 2003--2012 for validation, and 2013--2022 for testing. 
    As shown in Fig.~\ref{fig:shifts}, this split induces distribution shift, where the mean annual temperature of the test set is $0.90\,^{\circ}\mathrm{C}$ higher than the training set, corresponding to a phenological date shift (e.g., larch needle emergence occurs $6.22$ days earlier on average).

    \item \textbf{Annual temperature}
    Second, we rank years by their mean annual temperature. 
    We use the 15\% warmest years as test set, the next warmest 15\% as validation, and the remaining as training. 
    This creates a stronger shift with a $+1.23^{\circ}\mathrm{C}$ difference between training and testing, and an almost $8$ day advancement of larch needle emergence.

    \item \textbf{Elevation}
    For the third configuration, we split the data based on elevation to simulate stronger climate changes. We rank all site-years by elevation -- available in the site metadata -- and assign the highest 25\% to the test set, the next 15\% to validation, and the remaining to training. 
    This setting yields the most challenging shift among all splits: as shown in Fig.~\ref{fig:shifts}, both the temperature and phenology date distribution differ substantially, and reach values outside of the training range. 
\end{itemize}

\subsection{Implementation details}

All models are trained with MSE loss and optimized using Adam~\citep{kingma2014adam} with learning rate $10^{-4}$ and batch size 16. 
We use vector dimensions of $D=64$ and $F=128$. The discriminator $\mathbf{p}$ consists of a three-layer MLP with ReLU activations. 
For our method, we set $\tau=0.1$ in $\mathcal{L}_{\mathrm{rank}}$ and adopt the dynamic $\lambda$ scheduler from~\cite{ajakan2014DANN}. 
Models are trained for up to 300 epochs with model selection based on validation performance. Each experiment is repeated 10 times with different random initializations, and we report the mean results across runs. We evaluate performance using the coefficient of determination ($R^2$), root mean squared error (RMSE), and mean absolute error (MAE). All experiments are conducted on one NVIDIA GeForce RTX 3080 GPU.  
Additional implementation details are provided in the Supplementary Material (Sec.~S\ref{x_implement_details}) and in our \href{https://github.com/SherryJYC/MIRANDA}{official repository}.

\subsection{Results}

\begin{table*}[t]
\caption{
\textbf{Main results of three data splits.}
We compare \MethodName with PhenoFormer and five domain adaptation baselines across three data splits representing progressively stronger distribution shifts: Chronological, Annual Temperature, and Elevation. For reference, we also report the performance of the process-based model (M1), highlighted in blue. Evaluation metrics include $R^2$ ($\uparrow$), RMSE ($\downarrow$), and MAE ($\downarrow$). 
Results are averaged over $10$ repetitions and $5$ tree species, and we also report the standard deviation of $R^2$ in brackets.
}
\resizebox{\textwidth}{!}{
\begin{tabular}{lrrrrrrrrrrrrrrrrrr}
\toprule
    Dataset split $\rightarrow$ & \phantom{a} & \multicolumn{3}{c}{1. Chronological} & \phantom{a}& \multicolumn{3}{c}{2. Annual Temperature} & \phantom{a}&  \multicolumn{3}{c}{3. Elevation }  & \phantom{a} & Training \\ 
    
    Model  && \small $R^2$ $\uparrow$& \small RMSE $\downarrow$ & \small MAE $\downarrow$ && \small$R^2$ $\uparrow$ & \small RMSE $\downarrow$ & \small MAE $\downarrow$ &&\small $R^2$ $\uparrow$  &\small RMSE $\downarrow$  &\small  MAE $\downarrow$  && time (min)\\ \cmidrule{1-1} \cmidrule{3-5} \cmidrule{7-9} \cmidrule{11-13} 
   \\

   PhenoFormer Vanilla \cite{garnot2025phenoformer} &&  0.50 (0.01)  & 8.96 & 6.82 && 0.41 (0.02)  & 9.56 & 7.26 && 0.17 (0.02)  & 11.67 & 9.00 && 15\\   
   +DANN \cite{ajakan2014DANN}&&  0.49 (0.01) & 9.08 & 6.98 && 0.38 (0.06)  & 9.78 & 7.44 && -0.20 (0.21)  &14.03 & 11.32 && 17\\
   +ADDA \cite{tzeng2017ADDA} &&  0.23 (0.15)  & 11.14 & 8.80 && 0.12 (0.10)  & 11.64 & 9.18 && -0.61 (0.16)  & 16.22 & 13.33 && 26 \\
   +CORAL \cite{CORAL, sun2016deepCORAL} &&  \textbf{0.51} (0.01) & \textbf{8.93} & \textbf{6.79} && 0.42 (0.02) & 9.50 & 7.22 && -0.16 (0.15)  & 13.77 & 10.98 && 18\\
   +DANL \cite{romijnders2019DANL} &&  0.48 (0.02) & 9.17 & 7.03 &&0.30 (0.03)  & 10.44 & 8.06 && -0.01 (0.10)  & 12.90 & 10.24 && 18\\
   +AdaBN \cite{li2016AdaBN} &&  0.49 (0.01) & 9.05 & 6.89 && 0.40 (0.04)  & 9.64 & 7.34 && -0.14 (0.09)  & 13.71 & 10.92 && 15\\
   \textbf{\MethodName} (ours) &&  \textbf{0.51} (0.01) & 8.97 & 6.80 && \textbf{0.43} (0.02) & \textbf{9.43} & \textbf{7.16} && \textbf{0.22} (0.04)  & \textbf{11.32} & \textbf{8.63}  && 26 \\
   \color{blue}M1 \cite{blumel2012M1} &&  \color{blue}0.51 (0.05) & \color{blue}8.64 & \color{blue}6.77 && \color{blue}0.45 (0.05)  & \color{blue}9.40 & \color{blue}7.04 && \color{blue}0.25 (0.15)  & \color{blue}11.56 & \color{blue}8.53 && 150\\
\bottomrule
\end{tabular}
}
\label{tab:main-res}
\end{table*}

\paragraph{Competing methods} We evaluate \MethodName against five other domain adaptation approaches and a process model. We compare against divergence-based (CORAL \citep{CORAL}), adversarial (DANN \citep{ajakan2014DANN}, ADDA \citep{tzeng2017ADDA}), and normalization-based approaches (DANL \citep{romijnders2019DANL}, AdaBN \citep{li2016AdaBN}). We also compare against a vanilla PhenoFormer without adaptation. For fairness the vanilla model is set with two transformer layers as in \MethodName, instead of one. Lastly, we evaluate the M1 model \cite{blumel2012M1} which was the strongest process-based model in a previous study using the same dataset \cite{garnot2025phenoformer}. 

\paragraph{Main results}
Overall, \MethodName achieves the strongest overall performance among deep learning methods. On the chronological and annual temperature splits (moderate shifts), \MethodName achieves competitive or superior results compared to all baselines. In the elevation split, the most challenging setting with substantial climatic and phenological distribution shift, \MethodName significantly outperforms prior adaptation methods. Compared to the vanilla PhenoFormer, \MethodName improves $R^2$ by approximately $1\%$ (chronological), $2\%$ (annual temperature), and $5\%$ (elevation). 
In contrast, existing adaptation methods degrade severely under elevation shift. For example, DANN drops from $R^2=0.49$ in chronological to $R^2=-0.20$ in elevation split. Similar instability is observed for ADDA and CORAL. These results suggest that conventional domain adaptation techniques, which primarily align final-layer representations or rely solely on normalization adjustments, are insufficient for phenology forecasting under climate change. 
The process-based M1 model outperforms all deep learning approaches including ours, showcasing the robustness of mechanistic models. In terms of $R^2$, \MethodName closes the performance gap on the chronological split, and reduces it to $2-3\%$ on the other two splits.  
Notably, our method also exhibits lower standard deviation across runs and species than M1 (e.g., 0.04 vs. 0.15 on the elevation split). This suggests that, although M1 can perform well in settings aligned with its mechanistic assumptions, its behavior is less consistent under unseen conditions. In contrast, \MethodName achieves comparable robustness with greater stability. This brings deep phenology models closer to the extrapolation properties traditionally associated with process-based models while maintaining the flexibility of data-driven modelling. For example process models require one model to be trained per species, while \MethodName trains a single model to predict the $5$ species' dates. As shown on the last column of Tab. \ref{tab:main-res} this results in a $6\times$ shorter training time. Additionally, while the M1 model is only based on temperature and day length features, \MethodName also takes into account the  pressure and precipitation variables. This did not improve performance on our dataset, but other potential drivers such as soil water content could be easily tested using our framework.

\begin{figure}[ht]
    \centering
    \includegraphics[width=\linewidth]{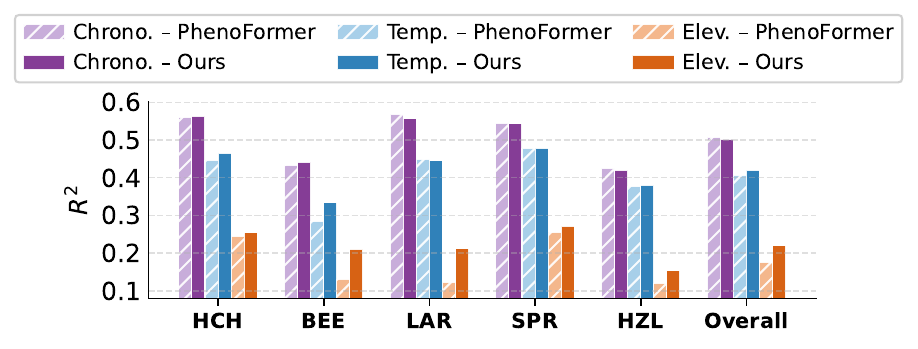}
    \caption{\textbf{Per-species performance} Histogram of $R^2$ per species for PhenoFormer and our model. Five species are: horse chestnut (HCH), European beech (BEE), European larch (LAR), common spruce (SPR), and hazel (HZL).}
    \label{fig:per_species_hist}
\end{figure}

\paragraph{Per-species performance}
We further examine the per-species performance of PhenoFormer and \MethodName in Fig. \ref{fig:per_species_hist}. 
Under the chronological split, the two models perform comparably across species. In the annual temperature split, the gains become more noticeable for European beech (BEE), and horse chestnut to a lesser extent. 
The most substantial differences appear in the elevation split, which represents the strongest climatic shift.  Here, \MethodName consistently outperforms PhenoFormer across all species, though the magnitude of improvement varies. For example, for horse chestnut (HCH), $R^2$ increases modestly from approximately 0.25 to 0.26, whereas for European larch (LAR), the improvement is much larger, rising from approximately 0.12 to 0.23. 
These variations highlight the fact that species respond differently to climatic distribution shifts. Across species and splits, our method is either on par or improves over the vanilla model.

\begin{figure}
    \centering
    \begin{subfigure}{.49\linewidth}
        \includegraphics[width=\linewidth]{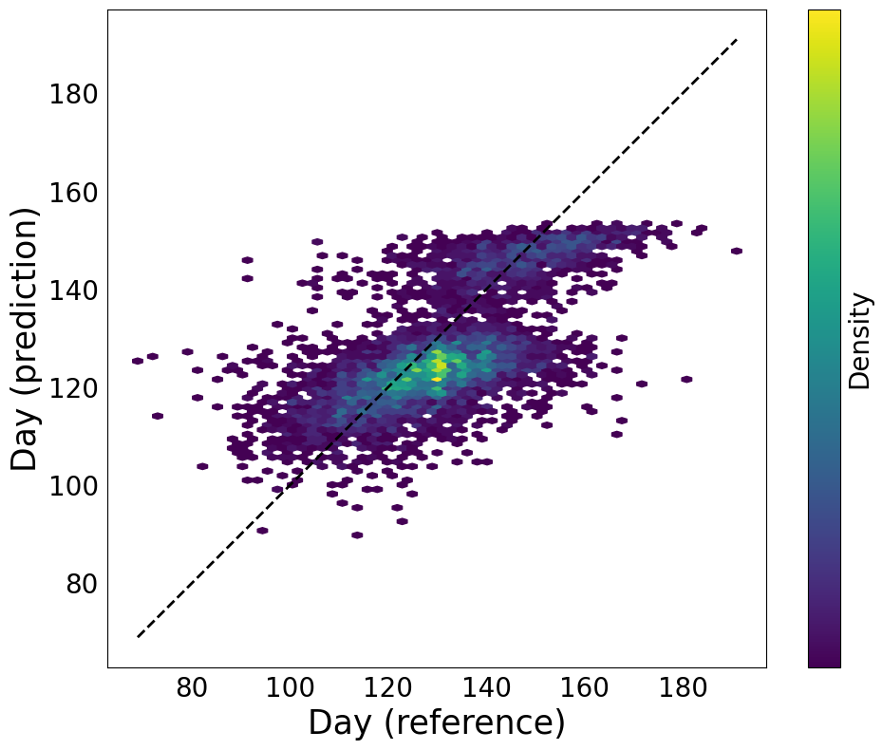}
    \end{subfigure} %
    \hfill
    \begin{subfigure}{.49\linewidth}
        \includegraphics[width=\linewidth]{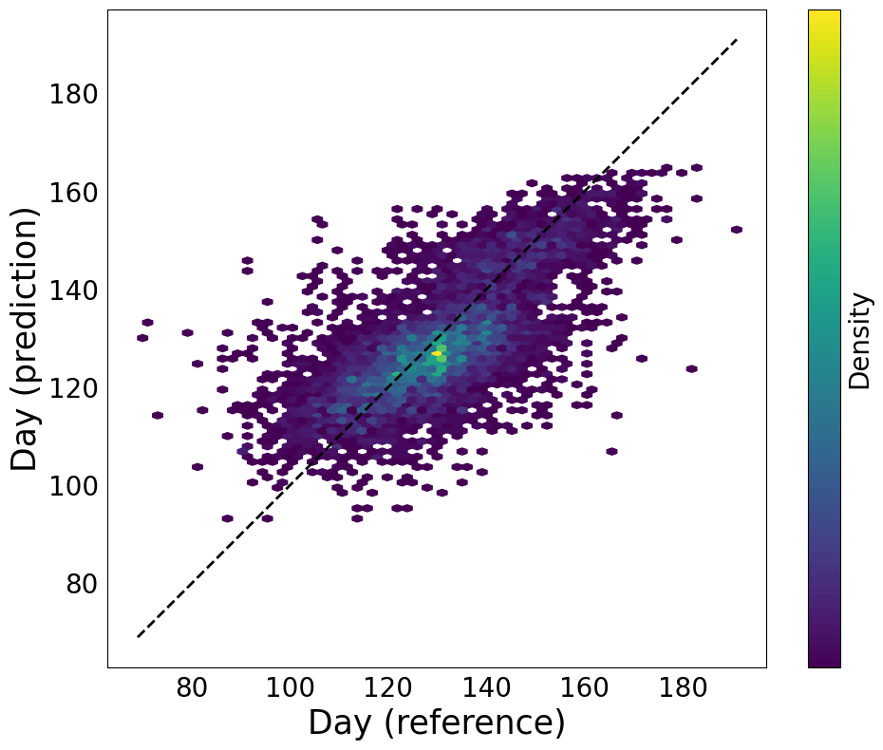}
    \end{subfigure} %
    \caption{Predicted (y-axis) vs true dates (x-axis) \textbf{before (left, with $R^2$ = 17\%) and after (right, with $R^2$ = 22\%) adaptation with MIRANDA on the elevation split.} Note that we plot the dates of the $5$ species together, which explains the different modes of the distribution. 
 }
    \label{fig:scatter_multispecies_highelve}
\end{figure}

\vspace{-11pt}
\paragraph{Qualitative results}

\begin{table*}[t!]
\caption{Ablations study of \MethodName. Starting from the PhenoFormer backbone and DANN adversarial training, we progressively introduce (1) mid-level feature alignment (mid-feat), (2) rank-based adversarial objective $\mathcal{L}_{\mathrm{rank}}$, and (3) hybrid layer normalization ($h\mathrm{LN}$).
Results demonstrate that each component contributes to improved robustness under domain shift, 
with the largest gains observed under the challenging Elevation split.}
\centering
\footnotesize{
\begin{tabular}{lrrrrrrrrrrrrr}
\toprule
    Dataset split $\rightarrow$ & \phantom{a} & \multicolumn{3}{c}{1. Chronological} & \phantom{a}& \multicolumn{3}{c}{2. Annual Temperature} & \phantom{a}&  \multicolumn{3}{c}{3. Elevation } \\ 
    
     && \small $R^2$ $\uparrow$& \small RMSE $\downarrow$ & \small MAE $\downarrow$ && \small$R^2$ $\uparrow$ & \small RMSE $\downarrow$ & \small MAE $\downarrow$ &&\small $R^2$ $\uparrow$  &\small RMSE $\downarrow$  &\small  MAE $\downarrow$  \\ \cmidrule{1-1} \cmidrule{3-5} \cmidrule{7-9} \cmidrule{11-13} 
   PhenoFormer Vanilla &&  0.50  & 8.96 & 6.82 && 0.41  & 9.56 & 7.26 && 0.17  & 11.67 & 9.00 \\ 
    +DANN \cite{ajakan2014DANN}&&  0.49  & 9.08 & 6.98 && 0.38  & 9.78 & 7.44 && -0.20  &14.03 & 11.32\\ \midrule
    \cmark mid-feat  &&  0.51  & 8.96 & 6.82 && 0.40  & 9.59 & 7.23 && 0.12  & 12.04 & 9.38 \\
   \cmark mid-feat \cmark $\mathcal{L}_{\mathrm{rank}}$  &&  0.50  & 8.97 & 6.82 && 0.42  & 9.45 & 7.18 && 0.19  & 11.54 & 8.87 \\
    \cmark mid-feat \cmark $h\mathrm{LN}$  &&  0.50  & 8.97 & 6.81 && 0.42  & 9.51 & 7.22 && 0.18  & 11.66 & 8.97 \\
   \textbf{\MethodName} &&  0.51 & 8.97 & 6.80 && 0.43  & 9.43 & 7.16 && 0.22  & 11.32 & 8.63 \\
\bottomrule

\end{tabular}
}
\label{tab:ablations}
\end{table*}
Fig.~\ref{fig:scatter_multispecies_highelve} provides a qualitative comparison of prediction distributions before and after adaptation in the elevation split.
Without \MethodName, the vanilla model's prediction saturate and struggle to predict phenological dates later than day of year $150$ leading to systematic underestimation. In comparison, the predictions of \MethodName do not seem to saturate and display an improved correlation with reference dates. This illustrates how our method successfully addresses the label shift induced by climate change on phenology modelling.

\subsection{Ablation study}

Starting from the PhenoFormer backbone with DANN~\citep{ajakan2014DANN} adversarial training as the baseline, we progressively introduce the three components of our method: 1) mid-level feature alignment (mid-feat) with binary domain classification as adversarial task, 2) rank-based adversarial training $\mathcal{L}_{\mathrm{rank}}$, 3) hybrid layer normalization ($h\mathrm{LN}$). 
Our method combines all three components. Results are reported in Tab.~\ref{tab:ablations}.
\paragraph{Mid-level feature alignment is critical}
Replacing late feature adversarial alignment (DANN) with mid-level feature alignment (\cmark mid-feat) is the single most important improvement. 
This suggests that aligning intermediate representations is more effective than aligning final high-level features when facing severe climatic distribution shift.
Conventional adversarial adaptation methods largely assume covariate shift, where the input distribution changes while the label distribution remains stable. Under this assumption, aligning final high-level features is sufficient. However, in phenology forecasting, domain shift also manifests as label shift (Fig.~\ref{fig:shifts}). Forcing alignment at the final representation level may thus be counterproductive and limit prediction range. Instead, enforcing domain-invariance only on intermediate features seems to strike a good balance to both address the climatic covariate shift, while also leaving space for domain-specific variations. 
This helps explain why domain adaptation methods that focus solely on high-level feature alignment perform poorly in Tab.~\ref{tab:main-res}.
\paragraph{Rank-based adversarial objective and hybrid layer normalization}
The rank-based adversarial loss $\mathcal{L}_{\mathrm{rank}}$ improves performance beyond standard binary adversarial training. By providing structured supervision for domain-invariant features, it yields more informative mid-level alignment, leading to consistent gains across splits. The effect is most pronounced under the elevation split, where $R^2$ increases from 0.12 to 0.19, surpassing the baseline PhenoFormer.
Adding the hybrid normalization layer ($h\mathrm{LN}$) also enhances robustness. By applying source-domain normalization statistics to both domains, $h\mathrm{LN}$ enforces consistent feature scaling while preserving domain-specific variations.

\section{Conclusion}
In this work, we addressed climatic distribution shift in deep phenology modelling through a domain adaptation framework tailored to this setting. Across multiple distribution shift settings, our method outperformed existing adversarial, divergence-based, and normalization-based adaptation approaches. We showed that applying adversarial regularization to intermediate features is key to handling both climate covariate and phenological label shift. Our hybrid normalization layer further improved the model’s ability to predict phenological dates outside the range observed in the source domain. Finally, we demonstrated that a rank-based adversarial objective encouraging year-invariant representations is better suited to phenology forecasting than conventional binary domain classification.
Altogether, these contributions narrow the robustness gap between deep learning and process-based models, which have traditionally demonstrated stronger extrapolation under shifted climatic conditions. These findings encourage further exploration of this problem in order to achieve even stronger data-driven performance with for example the inclusion of additional drivers, or exploring Fourier-based temporal features for improved robustness. 
By enhancing stability and transferability of data-driven phenology models, our work opens the way for a new set of tools for climate change-robust ecological forecasting. 

{
    \small
    \balance
    \bibliographystyle{ieeenat_fullname}
    \bibliography{phenocast}
}

\clearpage
\setcounter{page}{1}
\setcounter{section}{0}
\setcounter{figure}{0}
\maketitlesupplementary

\section{Implementation details}\label{x_implement_details}
We follow the implementation of PhenoFormer~\citep{garnot2025phenoformer} as our backbone and use two transformer encoder layers ($t_1$ and $t_2$). All models are trained using the MSE loss and optimized with Adam~\citep{kingma2014adam} with a learning rate of $10^{-4}$, batch size of 16, and up to 300 epochs. All domain adaptation methods adopt PhenoFormer as the backbone. For approaches involving adversarial training, the discriminator $p$ is implemented as a three-layer MLP with ReLU activations. Model-specific implementation details are provided below.

\begin{itemize}
    \item \textbf{DANN}~\citep{ajakan2014DANN}: 
    We adopt Phenoformer as the backbone and attach a domain discriminator to the high-level features extracted from the last transformer encoder layer. 
    The discriminator is trained with the standard binary domain classification objective, while the feature extractor is optimized adversarially via a gradient reversal layer, following the original DANN formulation.

    \item \textbf{ADDA}~\citep{tzeng2017ADDA}: 
    We initialize ADDA with the pretrained Phenoformer and follow the same experimental protocol as the other adaptation methods. 
    Specifically, we fine-tune the encoder and train a domain discriminator to align feature representations in an adversarial manner.

    \item \textbf{CORAL}~\citep{CORAL}: 
    We apply the CORAL loss to the high-level features extracted from the last transformer encoder layer, prior to the regression head, using the original formulation of the CORAL objective. 
    For a fair comparison, we set the CORAL weighting coefficient $\lambda$ to be the same as that used for our $\mathcal{L}_{\mathrm{rank}}$.

    \item \textbf{DANL}~\citep{romijnders2019DANL}: 
    The original DANL method introduces domain-agnostic normalization for batch normalization layers in convolutional architectures under the DANN framework. 
    Since Phenoformer is transformer-based and relies on layer normalization, we adapt DANL by replacing the standard layer normalization layers in the transformer encoder with our domain-agnostic layer normalization, while keeping the rest of the DANN training procedure unchanged.

    \item \textbf{AdaBN}~\citep{li2016AdaBN}: 
    AdaBN was originally proposed for models with batch normalization, where target-domain statistics are used during inference. 
    To incorporate AdaBN into Phenoformer, we insert a BatchNorm1d layer before the linear decoder and update its running statistics using unlabeled target samples during adaptation.

    \item \textbf{M1} \citep{blumel2012M1}:
    We use the implementation of M1 in the R package, Phenor \cite{hufkens2018phenor} and follow the same implementation setup specified in \cite{garnot2025phenoformer}.

    \item \textbf{\MethodName}:
    We set $\tau=0.1$ in $\mathcal{L}_{\mathrm{rank}}$ and adopt the dynamic $\lambda$ scheduler from~\cite{ajakan2014DANN}. For the embeddings used for rank loss, we set the dimension to 128.

\end{itemize}

\section{Additional experiment results}

\subsection{Guidance for rank-based adversarial training}

We investigate difference guidance for rank-based adversarial training: besides using year information as guidance, we explore the effect of using annual temperature and elevation as the guidance.

\begin{table}[h]
\caption{Different guidance for rank-based adversarial training under the Structured Temporal split.}
\centering
\begin{tabular}{lccc}
\toprule
Guidance & $R^2 \uparrow$ & RMSE $\downarrow$ & MAE $\downarrow$ \\
\midrule
Annual temperature & 0.50 & 9.01 & 6.85 \\
Elevation & 0.50 & 9.01 & 6.89 \\
Year (ours) & \textbf{0.51} & \textbf{8.97} & \textbf{6.80} \\
\bottomrule
\end{tabular}
\label{tab:guidance_structured}
\end{table}

\begin{table}[h]
\caption{Different guidance for rank-based adversarial training under the Annual Temperature split.}
\centering
\begin{tabular}{lccc}
\toprule
Guidance & $R^2 \uparrow$ & RMSE $\downarrow$ & MAE $\downarrow$ \\
\midrule
Annual temperature & 0.43 & 9.44 & 7.18 \\
Elevation & 0.42 & 9.44 & 7.19 \\
Year (ours) & \textbf{0.43} & \textbf{9.43} & \textbf{7.16} \\
\bottomrule
\end{tabular}
\label{tab:guidance_hotyear}
\end{table}

\begin{table}[h]
\caption{Different guidance for rank-based adversarial training under the Elevation split.}
\centering
\begin{tabular}{lccc}
\toprule
Guidance & $R^2 \uparrow$ & RMSE $\downarrow$ & MAE $\downarrow$ \\
\midrule
Annual temperature & 0.22 & 11.36 & 8.67 \\
Elevation & 0.19 & 11.55 & 8.86 \\
Year (ours) & \textbf{0.22} & \textbf{11.32} & \textbf{8.63} \\
\bottomrule
\end{tabular}
\label{tab:guidance_elevation}
\end{table}

\subsection{Additional qualitative results}

\begin{figure*}
    \centering
    \begin{subfigure}{.4\linewidth}
        \includegraphics[width=\linewidth]{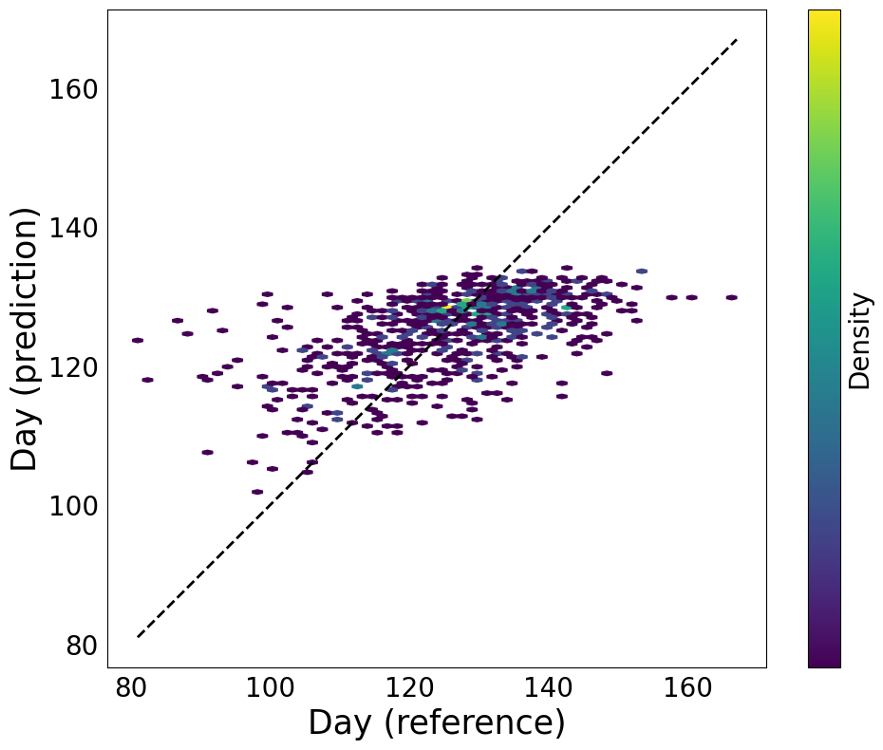}
    \end{subfigure} %
    \hfill
    \begin{subfigure}{.4\linewidth}
        \includegraphics[width=\linewidth]{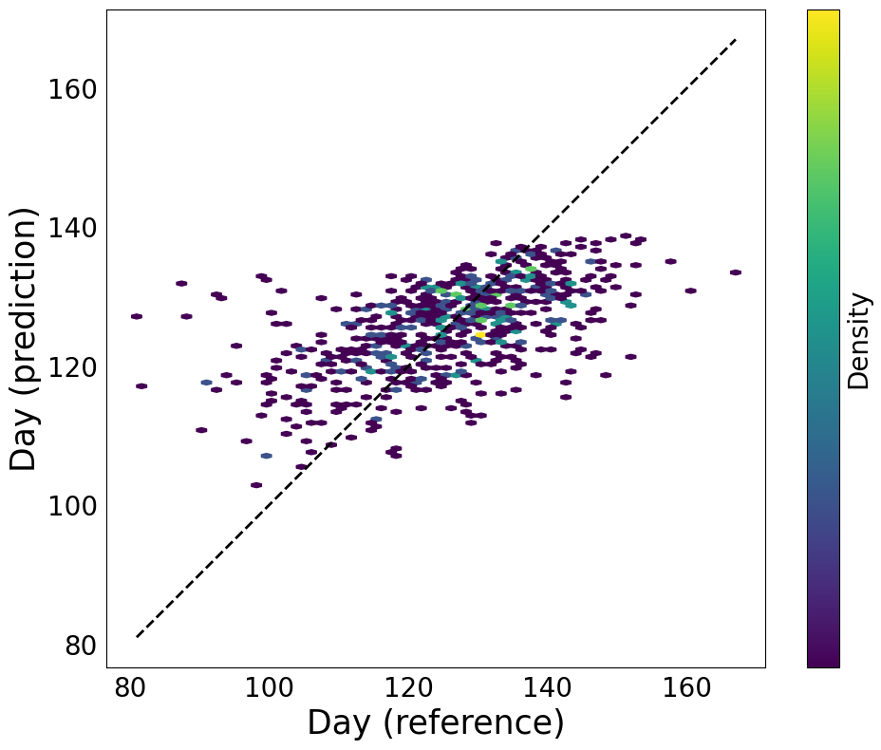}
    \end{subfigure} %
    \caption{Scatter plot of predicted (y-axis) vs true dates (x-axis) \textbf{before (left, with $R^2$ = 24\%) and after (right, with $R^2$ = 26\%) adaptation} for phenology task leaf unholding of horse chestnut in elevation data split.
 }
    \label{fig:scatter_ss1}
\end{figure*}

\begin{figure*}
    \centering
    \begin{subfigure}{.4\linewidth}
        \includegraphics[width=\linewidth]{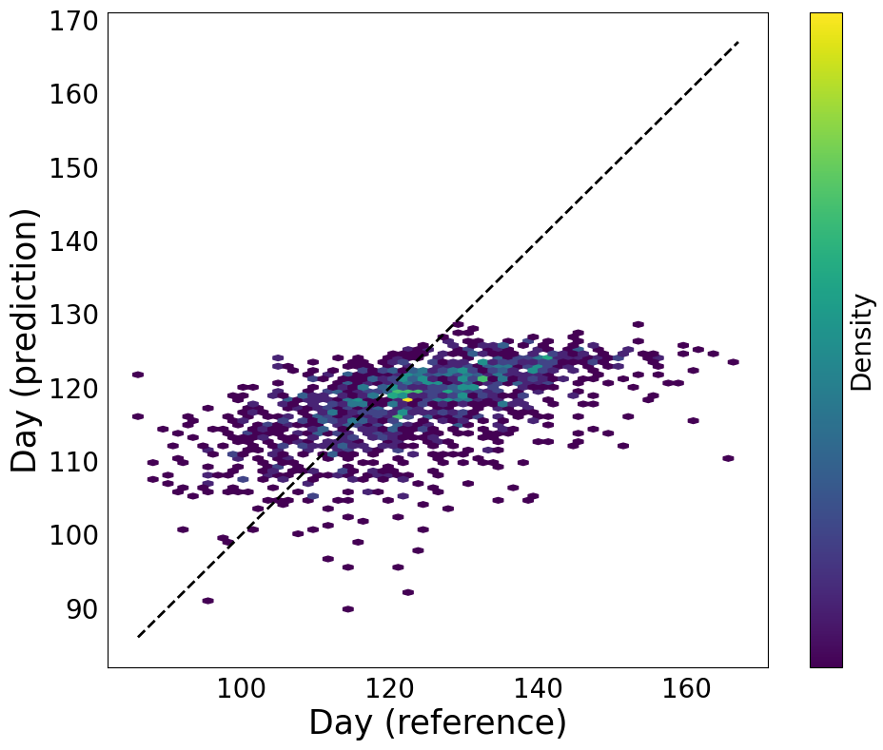}
    \end{subfigure} %
    \hfill
    \begin{subfigure}{.4\linewidth}
        \includegraphics[width=\linewidth]{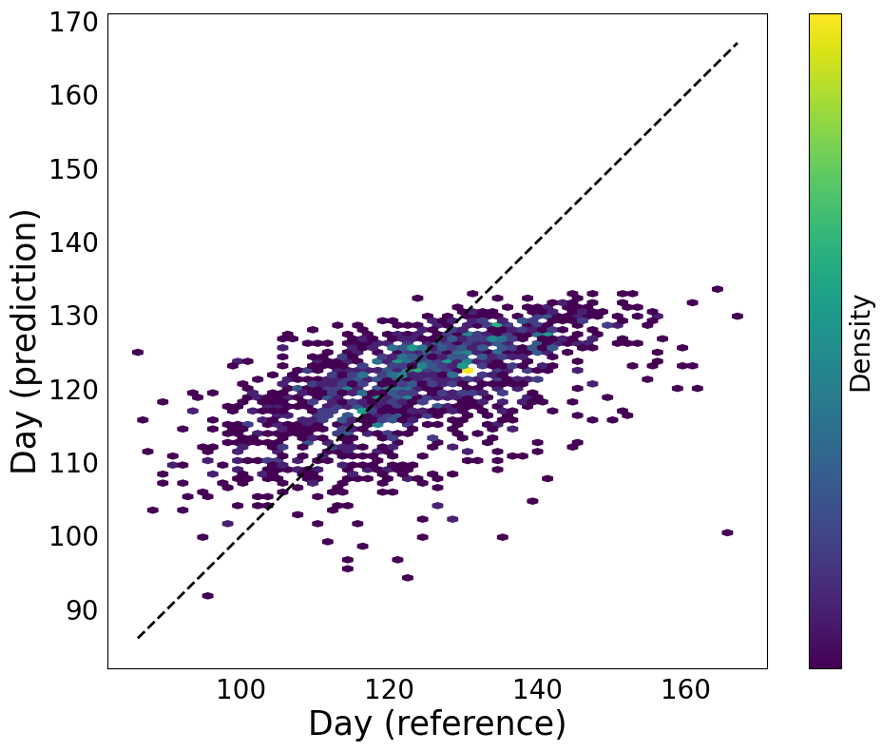}
    \end{subfigure} %
    \caption{Scatter plot of predicted (y-axis) vs true dates (x-axis) \textbf{before (left, with $R^2$ = 12\%) and after (right, with $R^2$ = 23\%) adaptation} for phenology task needle emergence of European larch in elevation data split.
 }
    \label{fig:scatter_ss2}
\end{figure*}

\begin{figure*}
    \centering
    \begin{subfigure}{.4\linewidth}
        \includegraphics[width=\linewidth]{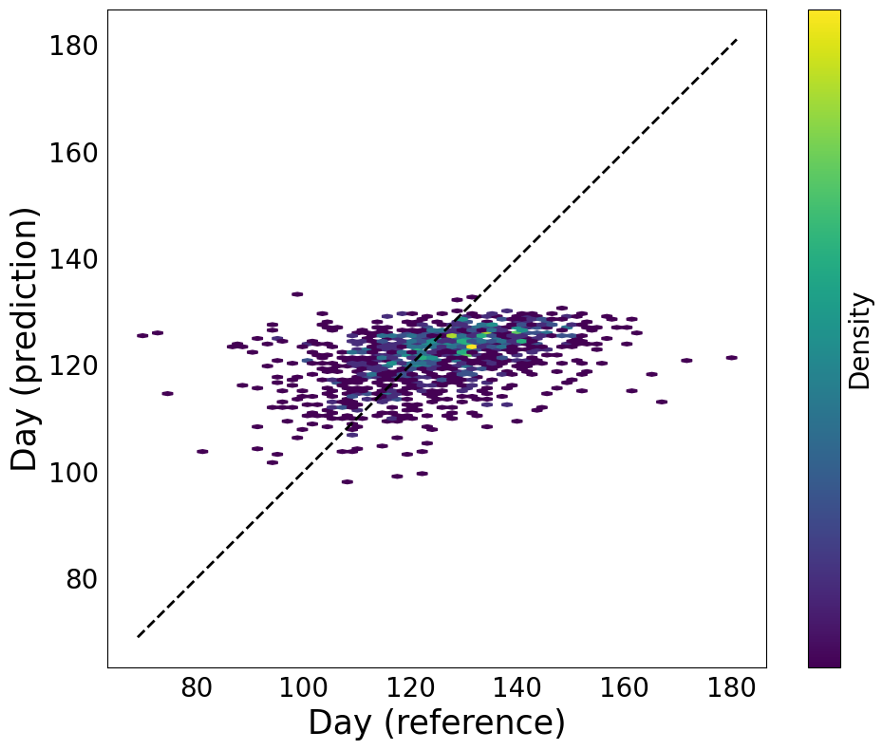}
    \end{subfigure} %
    \hfill
    \begin{subfigure}{.4\linewidth}
        \includegraphics[width=\linewidth]{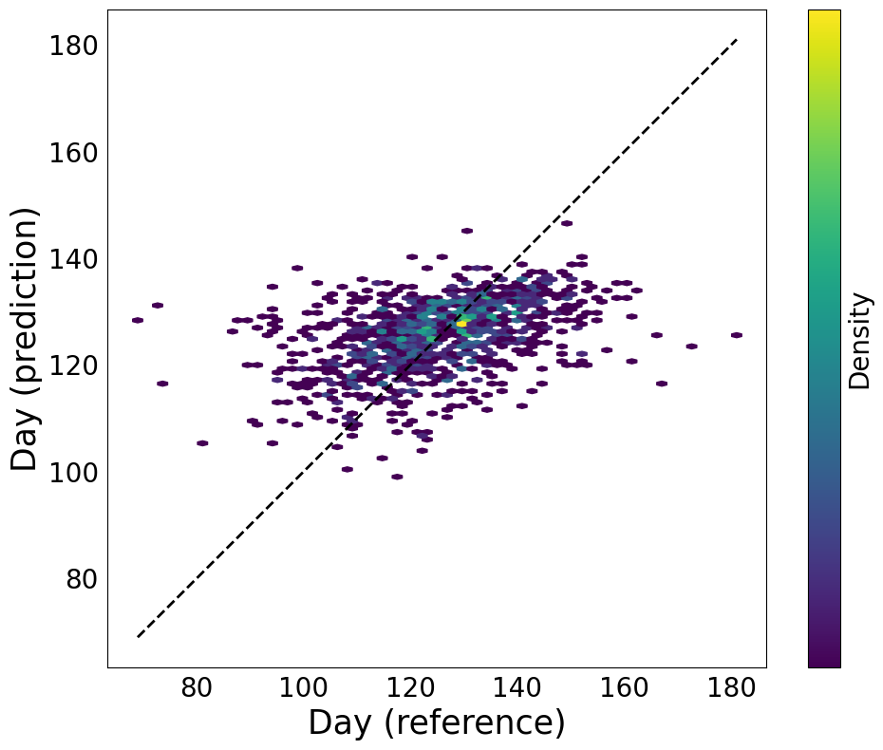}
    \end{subfigure} %
    \caption{Scatter plot of predicted (y-axis) vs true dates (x-axis) \textbf{before (left, with $R^2$ = 12\%) and after (right, with $R^2$ = 15\%) adaptation} for phenology task leaf unholding of hazel in elevation data split.
 }
    \label{fig:scatter_ss3}
\end{figure*}

\begin{figure*}
    \centering
    \begin{subfigure}{.4\linewidth}
        \includegraphics[width=\linewidth]{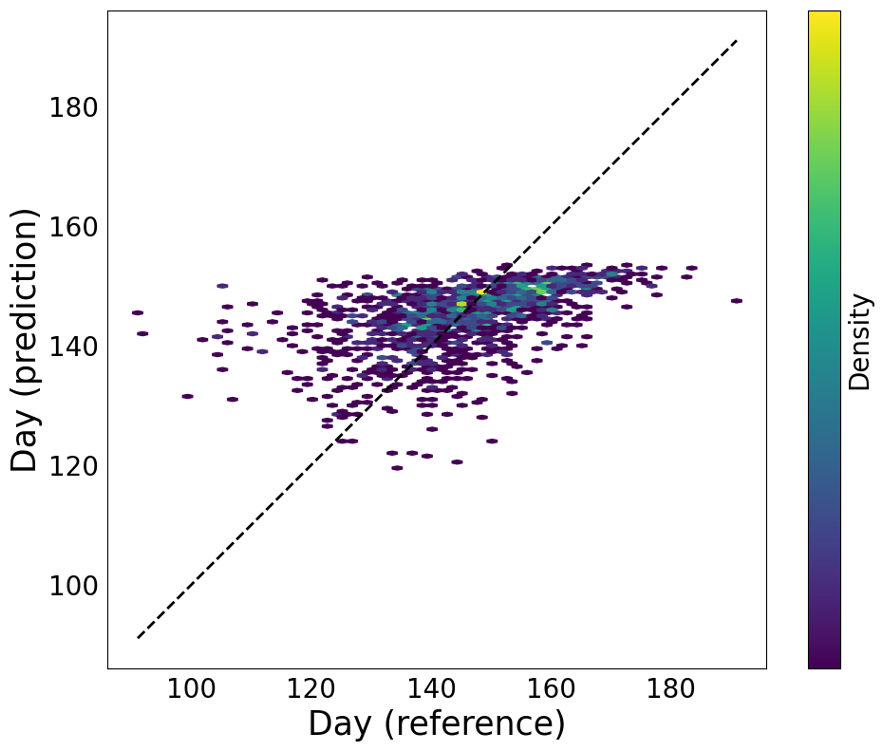}
    \end{subfigure} %
    \hfill
    \begin{subfigure}{.4\linewidth}
        \includegraphics[width=\linewidth]{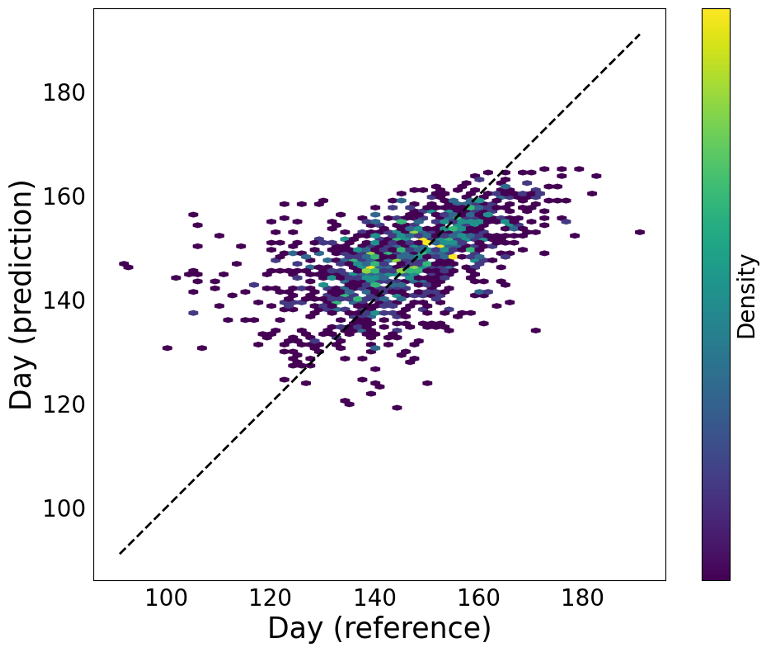}
    \end{subfigure} %
    \caption{Scatter plot of predicted (y-axis) vs true dates (x-axis) \textbf{before (left, with $R^2$ = 25\%) and after (right, with $R^2$ = 28\%) adaptation} for phenology task needle emergence of common spruce in elevation data split.
 }
    \label{fig:scatter_ss4}
\end{figure*}

\begin{figure*}
    \centering
    \begin{subfigure}{.4\linewidth}
        \includegraphics[width=\linewidth]{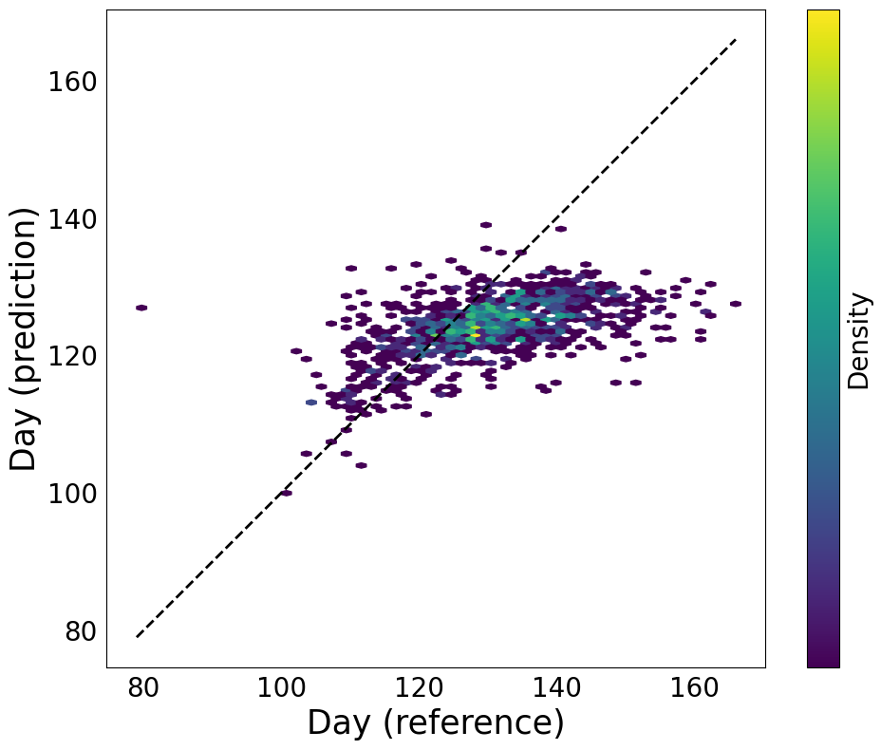}
    \end{subfigure} 
    \hfill
    \begin{subfigure}{.4\linewidth}
        \includegraphics[width=\linewidth]{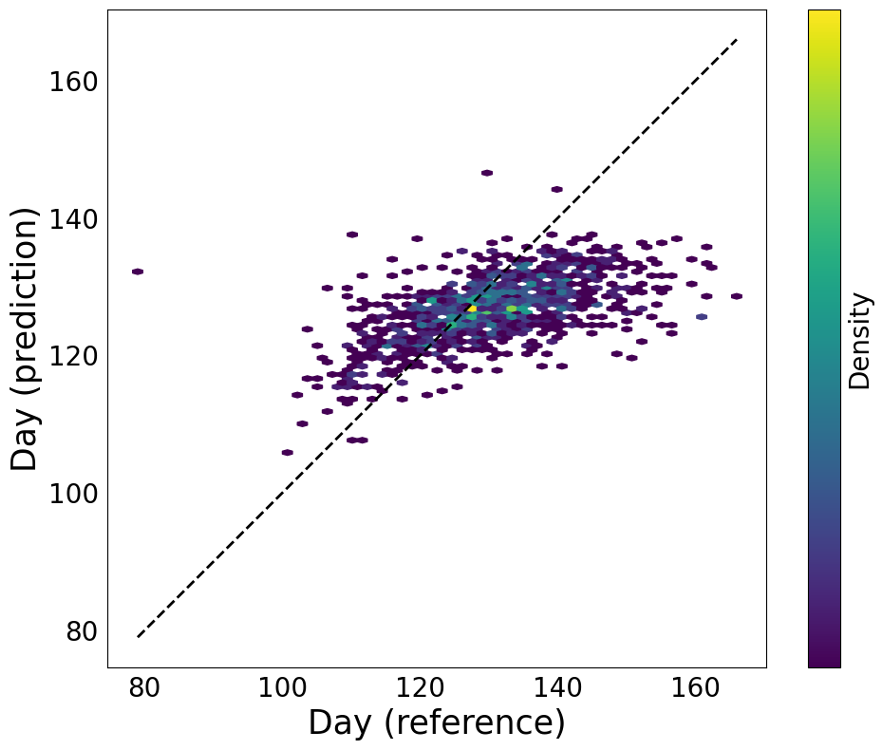}
    \end{subfigure} 
    \caption{Scatter plot of predicted (y-axis) vs true dates (x-axis) \textbf{before (left, with $R^2$ = 13\%) and after (right, with $R^2$ = 20\%) adaptation} for phenology task leaf unfolding of European beech in elevation data split.
 }
    \label{fig:scatter_ss5}
\end{figure*}

Detailed per-species scatter plots of PhenoFormer and our method of elevation data split are shown in Figure \ref{fig:scatter_ss1} to \ref{fig:scatter_ss5} .

\end{document}